\documentclass{article}
\usepackage[utf8]{inputenc}
\usepackage{spconf,amsmath,graphicx}
\usepackage{multirow}
\usepackage{diagbox}
\usepackage{caption}
\usepackage{hyperref}
\hypersetup{colorlinks,urlcolor=blue,linkcolor=blue,citecolor=blue}
\usepackage{algorithm}
\usepackage{algpseudocode}
\usepackage{makecell}
\usepackage{amsfonts}
\usepackage{booktabs}
\usepackage[style=ieee,maxcitenames=2, mincitenames=1, uniquelist=false,backend=bibtex]{biblatex}

\title{ArtiFact: A Large-Scale Dataset with Artificial and Factual Images for Generalizable and Robust Synthetic Image Detection}
\name{
\begin{tabular}{c}
Md Awsafur Rahman${}^{\text{\textsection}, 1}$, Bishmoy Paul ${}^{\text{\textsection}, 1}$, Najibul Haque Sarker ${}^{\text{\textsection}, 2}$, Zaber Ibn Abdul Hakim ${}^{\text{\textsection}, 2}$\\
Shaikh Anowarul Fattah ${}^{1}$ \end{tabular}}
\address{${}^1$ Dept. of EEE, BUET, Bangladesh\\
         ${}^2$ Dept. of CSE, BUET, Bangladesh}

\begin{document}

\maketitle

\begingroup\renewcommand\thefootnote{\textsection}
\footnotetext{Equal contribution}
\endgroup
\begin{abstract}
Synthetic image generation has opened up new  opportunities but has also created threats in regard to privacy, authenticity, and security. Detecting fake images is of paramount importance to prevent illegal activities, and previous research has shown that generative models leave unique patterns in their synthetic images that can be exploited to detect them. However, the fundamental problem of generalization remains, as even state-of-the-art detectors encounter difficulty when facing generators never seen during training. To assess the generalizability and robustness of synthetic image detectors in the face of real-world impairments, this paper presents a large-scale dataset\footnote{The dataset is available at \href{https://github.com/awsaf49/artifact}{https://github.com/awsaf49/artifact}} named ArtiFact, comprising diverse generators, object categories, and real-world challenges. Moreover, the proposed multi-class classification scheme, combined with a filter stride reduction strategy addresses social platform impairments and effectively detects synthetic images from both seen and unseen generators. The proposed solution significantly outperforms other top teams by 8.34\% on Test 1, 1.26\% on Test 2, and 15.08\% on Test 3 in the IEEE VIP Cup challenge at ICIP 2022, as measured by the accuracy metric.

\end{abstract}

\section{Introduction}


With the advent of deep learning technologies, an array of new methods have been introduced for synthetic image generation. These improvements have opened up new and exciting opportunities in creative arts, the entertainment industry, and advertising. But they are also posing a threat in regard to privacy, authenticity, and security—for example, generating fake images of a subject in different contexts.

To prevent these types of illegal and detrimental activities, developing technologies to detect these fake images have paramount importance. Most generative architectures have unique patterns in their synthetic images, which are absent in real images and vary with both the generator architecture and the dataset it is trained with \cite{marraFP}. 
Recent works on synthetic image detection exploited these artifacts present in generators utilizing color band correlation, intensity, Fourier spectra characteristics \cite{colorComp, freq, ganSAT, detectDM}. For the detection of these patterns and artifacts,  methods based on hand-crafted features and frequency domain analysis are outperformed by very deep CNN models~\cite{marra2018detection}.


The most fundamental problem that still remains is generalization, where even state-of-the-art detectors encounter difficulty when facing generators never seen during training \cite{detectDM}. With the rapid development of sophisticated generative models, it is impossible to include all possible variants in a detector's training set. This is further compounded by the prevalence of image impairment in compressed and resized social media images, which are particularly vulnerable to synthetic image-related fraudulent activities. The development of a viable solution is hindered by the lack of a proper benchmark dataset as existing relevant datasets are limited by a lack of diversity among generator sources and object classes. This necessitates the creation of a dataset featuring both real and fake images from diverse generators and object categories, with real-world challenges.

Given the foregoing context, this study provides the following significant contributions: \textbf{1)} A large-scale dataset namely ArtiFact is proposed, replete with diverse generators, object categories, and real-world impairments, poised to assess the efficacy of synthetic image detectors across a vast spectrum of sources and categories. \textbf{2)} To address the generalizability problem of detecting fake images from previously unseen generators, while also addressing real-world impairments that can impact the robustness of image detectors, a multi-class classification scheme along with filter stride reduction strategy is proposed for both generalizable and robust synthetic image detection.


\section{Related Work}
Deep learning-based generative models have revolutionized image synthesis in recent years. After the breakthrough in image synthesis by generative adversarial networks(GANs), the basic GAN framework has been extended for more diverse generations~\cite{gan_inversion, survey_on_gans}. Recent diffusion models~\cite{diffusion_survey} have shown impressive results in generating high-quality images with realistic textures and details.

Generative models have unique patterns that can be used to attribute them~\cite{yuFP}. Despite ongoing efforts to reduce these patterns in generators, even recent proposals such as Diffusion Models are not 
 free from them~\cite{detectDM}. 
Color band inconsistencies~\cite{colorComp} or lack of variation in color intensity~\cite{ganSAT} can act as unique identifiers for generative models. Fourier domain analysis also reveals unique signatures of generative models that can be used for model attribution~\cite{freq}.
Marra et al.~\cite{marra2018detection} shows that pre-trained state-of-the-art CNN image classification models outperform CNN models specifically designed from scratch for this task. Recent works on synthetic model attribution utilize supervised training of pre-trained CNN models with augmented data ~\cite{easytospot, gragnaniello2021gan}. Although they have shown promising results in detecting generators whose images are included in the training dataset, these methods have difficulty generalizing to unseen generators and also are susceptible to image impairments.



Existing datasets for this task are either limited in the number of diverse generators or object categories. For example, a dataset comprising five classes, including one real and four synthetic classes, is introduced by~\cite{yuFP}. However, this dataset only includes GAN models without any accompanying image data. Bui et al.~\cite{repmix} introduces a dataset with ten classes, comprising eight generator classes and two real classes. But this dataset still suffers from a lack of diversity of generators and object categories. Therefore, the need for more diverse and comprehensive datasets for synthetic image detection persists.

\section{Methodology}

\subsection{Proposed ArtiFact Dataset} \label{ssec:externalDataset}

The challenge of evaluating the performance of synthetic image detectors in terms of their generalizability and robustness requires a comprehensive dataset that meets specific requirements. These requirements include 1) a diversity of generators, including GAN, Diffusion, fully manipulating, and partially manipulating, 2) a diversity of object categories, encompassing various types rather than a few types, and 3) reflecting real-world scenarios by incorporating impairments resulting from social platforms. However, current datasets lack these features, limiting the ability to evaluate the detectors fully, and providing only a partial view of their robustness and generalizability. To address this issue, a large-scale dataset namely \textbf{ArtiFact} (\textbf{Arti}ficial and \textbf{Fact}ual), has been proposed which integrates diverse generators, object categories, and real-world impairments, providing researchers with a more comprehensive understanding of synthetic image detectors' generalizability and robustness.

\begin{figure}[h]
    \centering
    \includegraphics[scale=0.120]{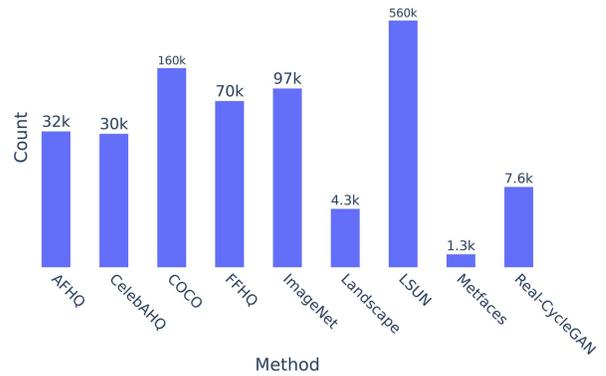}
    \caption{Distribution of different methods for real class}
    \label{fig:artifact-real}
    \vspace*{-0.4cm}
\end{figure}

\begin{figure}[h]
    \centering
    \includegraphics[scale=0.120]{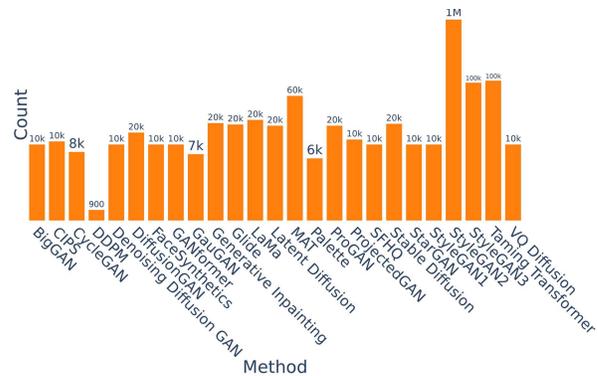}
    \caption{Distribution of different methods for fake class}
    \label{fig:artifact-fake}
    \vspace*{-0.4cm}
\end{figure}

\subsubsection{Dataset Characteristics}
To include a diverse collection of real images from multiple categories, including Human/Human Faces, Animal/Animal Faces, Places, Vehicles, Art, and many other real-life objects, the proposed dataset utilizes 8 sources~\cite{gan_inversion, stylegan3, cips, multimodal_survey} that are carefully chosen. Additionally, to inject diversity in terms of generators, the proposed dataset synthesizes images from 25 distinct methods~\cite{gan_inversion, survey_on_gans, diffusion_survey, stylegan3, glide, generative_inpainting, multimodal_survey, taming_xfr, cips, face-syn, stargan, mat,SFHQ, saharia2022palette}. Specifically, it includes 13 GANs, 7 Diffusion, and 5 other miscellaneous generators. On the other hand, in terms of syntheticity, there are 20 fully manipulating and 5 partially manipulating generators, thus providing a broad spectrum of diversity in terms of generators used. The distribution of real and fake data with different sources is shown in Fig.\ref{fig:artifact-real} and Fig.\ref{fig:artifact-fake}, respectively. The dataset contains a total of 2,496,738 images, comprising 964,989 real images and 1,531,749 fake images. The most frequently occurring categories in the dataset are Human/Human Faces, Animal/Animal Faces, Vehicles, Places, and Art.

\subsubsection{Dataset Methodology}
To ensure significant diversity across different sources, the real images of the dataset are randomly sampled from source datasets containing numerous categories, whereas synthetic images are generated within the same categories as the real images. Captions and image masks from the COCO dataset are utilized to generate images for text2image and inpainting generators, while normally distributed noise with different random seeds is used for noise2image generators. In both cases, the generator's default configuration is employed to produce images. To ensure that the proposed dataset reflects real-world scenarios, both real and synthetic images of the dataset undergo different impairments in accordance with the IEEE VIP Cup 2022 standards~\cite{detectDM}. Specifically, random cropping with a ratio of $r = \frac{5}{8}$ and minimum and maximum crop sizes of $160$ and $2048$, respectively, are applied to the images. They are then resized to $200 \times 200$ before being compressed using the JPEG format with quality $Q_f \in [65, 100]$.  Thereby, the proposed dataset accurately reflects real-world conditions, making it an ideal benchmark for evaluating the performance of synthetic image detectors.

\subsection{Proposed Detection Scheme}

\begin{figure}[t]
    \centering
    \hspace*{-0.2cm} 
    \includegraphics[scale=0.120]{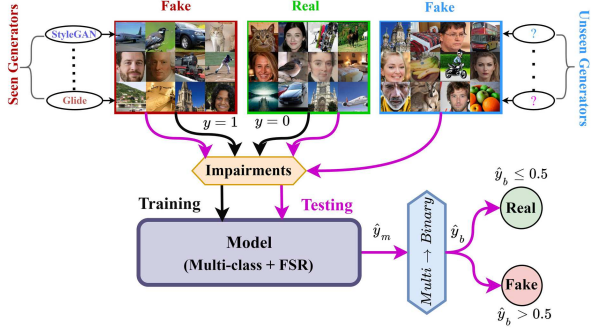}
    \caption{Visual Summary of proposed method}
    \label{fig:summary}
    \vspace*{-0.6cm}
\end{figure}

Current approaches struggle with both unseen generators and impairments. Therefore, a robust and generalizable detector is necessary which is 1) capable of generalizing for both seen and unseen generators, with or without impairments, and 2) do not lose critical information by preserving generator traces in presence of impairments. To meet these requirements, an effective method has been proposed utilizing Multi-class Scheme (Section~\ref{ssec:multiClass}) to tackle the first challenge and Filter Stride Reduction (FSR) (Section~\ref{ssec:fsr}) to handle the second challenge. As depicted in Fig.~\ref{fig:summary}, the proposed method utilizes real and fake images from seen generators for training and testing, with fake images from unseen generators used solely as test data for predicting binary levels of authenticity in multi-class settings.


\subsubsection{Multi-class Classification Scheme} \label{ssec:multiClass}
 
\begin{figure}[h]
    \centering
    \includegraphics[scale=0.120]{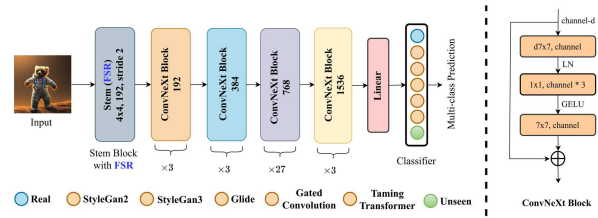}
    \caption{ConvNeXt backbone with filter stride reduced (FSR) stem block and multi-class head including extra unseen class}
    \label{fig:multi}
    \vspace*{-0.6cm}
\end{figure}

This paper introduces a novel multi-class classification scheme to enhance the generalization and robustness of synthetic image detection. As depicted in Figure~\ref{fig:multi}, the traditional binary classification problem of distinguishing real and fake images is transformed into a multi-class classification task with seven classes in accordance with IEEE VIP Cup~\cite{detectDM}. Specifically, the proposed approach includes one class for identifying real images, five classes for identifying fake images from five seen generators (generators whose synthesized images are present in the training dataset), and one unique class, namely Unseen Fake (UF) for identifying fake images from unseen generators (data from generators that are not present during training). Traditional approaches are inadequate to account for these UF images. However, when the proposed method encounters images from previously unseen generators, the learning from the diverse UF class aids in labeling them as fake. Besides, the multi-class approach leverages the notion that increasing the number of classes can significantly improve the performance~\cite{imagenet-21k} as it exposes the model to a diverse range of visual concepts and relationships between classes, leading to the learning of more generalized and robust features. The proposed method generates multi-class predictions but the evaluation metric expects binary-class, thus multi-class prediction is converted to binary-class taking the complement of real-class prediction.

\subsubsection{Filter Stride Reduction (FSR)}
\label{ssec:fsr}
The utilization of social media networks often results in the resizing and compression of images, which can result in the loss of critical information causing harm to invaluable traces of generator artifacts. This problem is exacerbated when the image is processed by modern CNNs and Vision Transformers, where the reduction of resolution in the stem block can further damage crucial information. To tackle this problem, a novel approach is put forth, which reduces the filter stride in the stem block of the ConvNeXt~\cite{convnext} backbone by $2 \times$ as shown in Fig.~\ref{fig:multi}. This approach enables the reduction of information loss and preservation of generator artifacts while keeping architectural integrity and utilizing pre-trained weights, thus resulting in substantial improvement in performance.


\section{Experiments}

\subsection{Experimental Setup}
The proposed approach employs ConvNext-Large~\cite{convnext} as the backbone and a resolution of $200 \times 200$, with the Adam optimizer and Exponentially Decay scheduler with an initial learning rate of $10^{-4}$. Furthermore, it utilizes Categorical Cross Entropy loss with label smoothing, with $\varepsilon = 5 \times 10^{-2}$.  To evaluate the detector’s generalizability reliably, a four-fold hybrid cross-validation scheme is implemented with balanced accuracy which applies KFold~\cite{sklearn} for real class and five seen fake classes where data from the same generators can appear in both the train and test sets as well as GroupKFold~\cite{sklearn} for the unseen fake class where there is no overlap between generators in the train and test sets. To prevent overfitting, various augmentations are also employed randomly, including scale-shift-rotate-shear, contrast-brightness-hue, flips, and cutout.


\subsection{Ablation Study}
An ablation study is conducted to evaluate the proposed method's effectiveness in terms of balanced accuracy, and the results are presented in Table \ref{tab:ablation}. The table shows that the Multi-class scheme with Filter Stride Reduction (FSR) and an Unseen Fake (UF) class achieved the highest accuracy of 87.62\%. This approach outperformed the traditional Binary-class method by 9.41\%, demonstrating the efficacy of the proposed method.

\begin{table}[h]
\centering
\caption{Ablation study of the proposed method.}
\captionsetup{justification=raggedright,singlelinecheck=false}
\label{tab:ablation}
\begin{tabular}{lccc} 
\toprule
Method               & Accuracy            \\ 
\midrule
Binary-class                 & 78.21           \\
Binary-class + FSR    & 81.30          \\ 
Multi-class & 83.12 \\
Multi-class + UF class & 84.98 \\
Multi-class + FSR & 85.56 \\
\midrule
\textbf{Multi-class + FSR + UF class} & \textbf{87.62}   \\
\bottomrule
\end{tabular}
\end{table}

\subsection{Result on IEEE VIP Cup 2022}
The performance of the proposed method is evaluated in IEEE VIP Cup~\cite{detectDM} competition at ICIP 2022 using a small portion of the proposed ArtiFact dataset, totaling 222K images of 71K real images and 151K fake images from only 13 generators. As shown in Table~\ref{tab:vip22}, the proposed method consistently outperforms other top teams on the leaderboard by a significant margin, with an improvement of 8.34\% on Test 1, 1.26\% on Test 2, and 15.08\% on Test 3, as measured by the accuracy metric, thus validating the efficacy of the proposed method. It is important to note that the Test data is kept confidential from all participating teams. Additionally, the generators used for the Test 1 data are known to all teams, whereas the generators for Test 2 and Test 3 are kept undisclosed. Given that all the teams sourced train data from various sources, there is a possibility of overlap between generators used in Test 2 and Test 3 data.

\begin{table}
\centering
\caption{Accuracy (\%) of Top3 Teams in IEEE VIP Cup 2022}
\captionsetup{justification=raggedright,singlelinecheck=false}
\label{tab:vip22}
\begin{tabular}{lccc} 
\toprule
Team Names               & Test 1            & Test 2           & Test 3            \\ 
\midrule
Sherlock                 & 87.70           & 77.52          & 73.45           \\
FAU Erlangen-Nürnberg    & 87.14           & 81.74          & 75.52           \\ 
\midrule
\textbf{Megatron (Ours)} & \textbf{96.04} & \textbf{83.00} & \textbf{90.60}  \\
\bottomrule
\end{tabular}
\end{table}

\subsection{Comparison with existing approaches}

A quantitative comparison of the proposed method with existing techniques has been presented in Table~\ref{tab:result_of_methods} which clearly demonstrates that the proposed method exhibits exceptional performance, surpassing other methods by a significant margin in terms of balanced accuracy. These findings conclusively show the effectiveness of the proposed method in the field of synthetic image detection.

\begin{table}[h]
\centering
\caption{Comparison of the proposed method with existing approaches.}
\captionsetup{justification=raggedright,singlelinecheck=false}
\label{tab:result_of_methods}
\begin{tabular}{lccc} 
\toprule
Method               & Accuracy            \\ 
\midrule
Joel et al. \cite{freq} & 63.19 \\
Francesco et al. \cite{marra2018detection} & 79.28 \\
Wang et al. \cite{easytospot}                 & 79.95           \\
Gragnaniello et al. \cite{gragnaniello2021gan}    & 81.63          \\ 
\midrule
\textbf{Multi-class + FSR + UF class (ours)} & \textbf{87.62}   \\
\bottomrule
\end{tabular}
\end{table}

\section{Conclusion}
In this study, a novel dataset consisting of 2.4M images for fake image detection containing diverse generators, and object categories with real-world scenarios has been presented, along with a robust methodology that is immune to social-platform impairments and attacks from unseen generators. The proposed multi-class scheme with a dedicated class for unseen generators demonstrates exceptional performance. Additionally, the proposed filter stride reduction effectively combats the loss of critical information caused by social-platform impairments. Thus, the proposed solution is a promising contender in the field of synthetic image detection, with the potential to address crucial forensic issues related to synthetic images.

\section{Acknowledgment}
The authors would like to express their gratitude to the IEEE Signal Processing Society, GRIP of the University Federico II of Naples (Italy), and NVIDIA (USA) for hosting the IEEE VIP Cup competition which acted as motivation for this work.


\printbibliography

\end{document}